\documentclass[10pt,journal,compsoc]{IEEEtran}

\usepackage{epsfig}
\usepackage{subfigure}
\usepackage{calc}
\usepackage{amssymb}
\usepackage{amstext}
\usepackage{amsmath}
\usepackage{amsthm}
\usepackage{multicol}
\usepackage{multirow}
\usepackage{pslatex}
\usepackage{color}

\begin{document}
\title{Analysing the Direction of Emotional  Influence in  Nonverbal Dyadic Communication:\\ A Facial-Expression Study}

\author{Maha Shadaydeh$^*$, Lea M\"uller$^*$, Dana Schneider$^*$, Martin 
Th\"ummel, Thomas Kessler  and Joachim 
Denzler, {\it member,  IEEE} 
\thanks{ $^*$ Co-first author;  corresponding author:  maha.shadaydeh@uni-jena.de}
\thanks{M. Shadaydeh, M. Th\"ummel,   and J. Denzler are with the Computer Vision Group, Department of Mathematics and Computer Science,  Friedrich Schiller University of Jena,  Ernst-Abbe-Platz 2, 07743 Jena, Germany.}
\thanks{L.  M\"uller is at the Max Planck Institute for Intelligent Systems, Max-Planck-Ring 4, 72076 T\"ubingen, Germany. This work was done during her master thesis at the computer vision group, FSU Jena, Germany.}
\thanks{ D. Schneider  and T.  Kessler  are at  the Department of Social Psychology, Friedrich Schiller University of Jena, Humboldtstrasse 26, 07743 Jena, Germany.}

}



\IEEEtitleabstractindextext{%
\begin{abstract}
Identifying the direction of emotional influence in a dyadic dialogue is of increasing interest in the psychological sciences with applications in psychotherapy, analysis of political interactions or  interpersonal conflict behavior.   Facial expressions are widely described as being automatic and thus hard to overtly influence. As such, they are a perfect measure for a better understanding of unintentional behavior cues about social-emotional cognitive processes. With this view, this study is concerned with the analysis of the direction of emotional influence in dyadic dialogue based on facial expressions only. We exploit computer vision capabilities along with causal inference theory for quantitative verification of hypotheses on the direction of emotional influence, i.e., causal effect relationships,   in dyadic dialogues.   We address two main issues. First, in  a  dyadic dialogue, emotional  influence occurs over  transient time intervals and with intensity and direction that are variant over time. To this end, we  propose  a relevant interval selection approach that we use prior to causal inference to identify those transient intervals where causal inference should be applied. Second, we propose to use fine-grained facial expressions that are present when strong distinct  facial emotions are not visible.  To specify the direction of influence,  we apply the concept of Granger  causality to the time series of facial expressions over selected relevant intervals. We tested our approach on  newly, experimentally obtained data.   Based on  quantitative verification of hypotheses on the direction of emotional influence, we were able to show that the proposed  approach is most promising to reveal the causal effect pattern  in various instructed interaction conditions. 
\end{abstract}

}

\maketitle

\IEEEdisplaynontitleabstractindextext

\IEEEpeerreviewmaketitle

\ifCLASSOPTIONcompsoc
\IEEEraisesectionheading{\section{Introduction}\label{sec:introduction}}
\else
\section{Introduction}
\label{sec:introduction}
\fi

\noindent Analysing the direction of influence between participants in dyadic interactions (i.e., human communication between two people)  is of increasing interest in the psychological sciences with applications in psychotherapy \cite{Tschacher2014},   critical political interactions \cite{Renoust2016}, and interpersonal conflict behavior \cite{Reeck2016},   among others. In any interpersonal communication,  attitudes towards the interacting partner play a crucial role for the outcome of this interaction (e.g., liking or disliking, engagement or disengagement,  trust or no trust towards the interacting partner) \cite{Ambady2000,Todorov2005,Fiske2007}. Often self-report measures or verbal cues are used for getting at attitudes towards the interacting partner. However, most interacting partners are very adapted towards socially desirable behavior and thus would not express negative attitudes in an overt manner. The concept of embodiment \cite{Tschacher2014,Gallese2005,Fuchs2009}    denotes the theoretical perspective that mental processes are not isolated from bodily processes.  With this view, psychology is becoming increasingly sensitized to investigate the close association between mental and bodily parameters.  In this paper, we are interested in the analysis of the direction of emotional influence in dyadic interactions using non-verbal facial expressions.

Facial expressions, especially basic emotion expressions, are widely described as being automatic, thus hard to overtly influence and independent of culture \cite{Ekman1976}. As such, they seem a perfect measure for a better understanding of unintentional behavioral cues about social-emotional cognitive processes.  Furthermore, there is recent evidence that non-verbal emotion processing is much more relevant than verbal emotion processing \cite{Filippi2017} and that language plays a marginal role in the perception of emotion \cite{Sauter2017}. 

Novel developments in computer vision yielded accurate,  open-source and  real-time tools to easily extract  bodily parameters from images and videos.  In this paper, a complete concept for analyzing the direction of emotional influence in a dyadic face to face dialogues, when starting with raw video material, is presented. We exploit computer vision capabilities for quantitative verification of hypotheses on the direction of emotional influence in experimentally recorded dyadic dialogues using facial expressions only, i.e.,  facial muscle activation (i.e., Action Units- AUs) \cite{ekman2002facial}. 

Human nonverbal communication is a process of continual two-sided influence. In a  dyadic interaction, such influence occurs over transient time intervals  with intensity and direction that are variant over time. To approach the problem of analysing the direction of influence,    we first present each person's facial emotional expressions  in the form of time series. To investigate who is influencing whom and how, we perform a  causal effect analysis on the obtained time series using  Granger causality (GC)  \cite{geweke1982measurement,GRANGER1980329},  the most widely used  method for causal inference  in diverse fields \cite{Peters2017,ClimateGC2013,Seth2015,Shadaydeh2019}. GC  is based on the idea that causes both precede and help predict their effects.  

The  contributions of this work can be summarized as follows. 
\begin{itemize}
\item[1.] The use of GC  for causal inference in nonverbal communication data has been addressed by several authors using different bodily parameters.  However,  to the best of our knowledge, no other work has used  GC  to identify the direction of emotional influence when it comes to facial expressions in dyadic dialogues. 
\item[2.] Using facial expressions in the upper and lower face regions,  we were able to extract  fine-grained facial features from the six basic emotions \cite{ekman1992argument} which can be present even when strong emotional expressions are not visible.
\item[3.] We present a relevant interval selection approach that we use prior to causal inference to identify those transient intervals where causal inference should be applied.  The  validation of the relevant interval selection on synthetic data  for improved causal inference has been presented  in  \cite{Mueller2019} along with initial results of this study on small data set. Here we show based on a larger real data set the superiority of such an approach in detecting the direction of emotional influence when compared to applying GC test on the entire time-series.
\item[4.]   The current work  is a study of   covert attitudes in interacting partners from a 2nd person perspective \cite{Schilbach2013}, meaning in a truly interactive manner. In the past studying real-time social encounters in an interactive manner was described as the “dark matter” of social (neuro-) science as it is typically explored from a 3rd person perspective \cite{Schilbach2013}. Most studies in the field of social cognition \cite{Happe2017} are conducted in a manner that interacting partners observe other interacting partners from a distance (e.g., by judging facial expressions of another person from a picture). However, our work goes a step into the the "dark matter" and is looking at interacting  partners that are readily involved, thus continuously responding to the others' actions. 

\end{itemize}

The remainder of this paper is organized as follows. Related work are summarized in Section \ref{sec:Related}. We introduce the experimental setup and used psychological hypotheses in Section \ref{sec:ExperimentalSetup}. The methodological details are presented in Section \ref{sec:methodology} followed by experimental results and discussion in Section \ref{sec:results}.  Finally,  Conclusions are given in Section \ref{sec:conclusion}.

\section{Related Work}
\label{sec:Related}
The topic of finding causal structures in nonverbal communication data is addressed by Kalimeri et al. \cite{kalimeri2012modeling}. In their paper, the authors used  GC  for modeling the effects that dominant people might induce when it comes to nonverbal behavior i.e.,   speech energy and body motion,  of other people. Besides audio cues, motion vectors and residual coding bit rate features from skin colored regions were extracted. In two systems, one for body movement and another one for speaking activity, with four-time series each, a small  GC  based causal network was used to identify the participants with high or low causal influence. Unlike our approach, the authors did not use facial expressions and did not identify relevant intervals in a  previous step, but used the entire time series instead. 

Kaliouby and Robinson \cite{el2005real} provided the first classification system for agreement and disagreement as well as other mental states based on nonverbal cues only. They used head motion and facial AUs together with a dynamic Bayesian network for classification. Also, a survey on cues, databases, and tools related to the detection of spontaneous agreement and disagreement was conducted  by Bousmalis et al. \cite{bousmalis2013towards}.  Despite their ingenious methods, these approaches did not investigate causal effect relations in a social interaction situation. 

Postma and Postma-Nilsenova \cite{Postma2016} used the convergent cross mapping method (a causal inference method originating from dynamical systems theory) to study nonverbal interactions. They concluded that there exists bidirectional causal coupling in facial dynamics. Unlike our study,  their method focuses on bidirectional causality only rather than identifying all possible directions of emotional influences. Further, their approach was applied to two time-series of single AU  over the entire time-series rather than a combination of two or more AUs presenting emotions in selected intervals.  Beyond this, Sheerman-Chase et al. \cite{sheerman2009feature} used visual cues to distinguish between states such as thinking, understanding, agreeing, and questioning to recognize agreement. Most intriguingly,  Matsuyama et al. \cite{matsuyama2016socially} developed a socially-aware robot assistant responding to visual and vocal cues. For visual features, the robot extracted facial cues (based on OpenFace \cite{baltrusaitis2018openface}) using landmarks, head pose, gaze, and facial AUs. Conversational strategies that build, maintain, or destroy connecting relationships were classified.  The researchers' approach investigates the building of a social relationship between a human and a robot; however, this study does not deal with the time-variant direction of causal effect relations.  

Joo et al. \cite{joo2019towards} recently presented a motion capture data set and a neural network architecture to analyze the direction of influence in a triadic interaction. Their approach focuses on predicting social signals, such as speaking status, social formation, and body gesture. The face is represented using  deformable face model with five facial expression parameters. In contrast to Joo et al., we are only interested in facial expressions. This requires a more detailed and supervised representation of facial motion.

Several methods,  based on temporal causality analysis, have been proposed for categorizing interactions of individuals in video sequences  \cite{Prabhakar2010,Prabhakar2012,Ayazoglu2013}.  These methods use Granger causality,  but they are based on motion analysis of different individuals and are not concerned with facial emotional influences.   

\section{Experimental Setup and Psychological Hypotheses}
\label{sec:ExperimentalSetup}
\begin{figure}[!h]
\centering
 \includegraphics[width=0.4\textwidth]{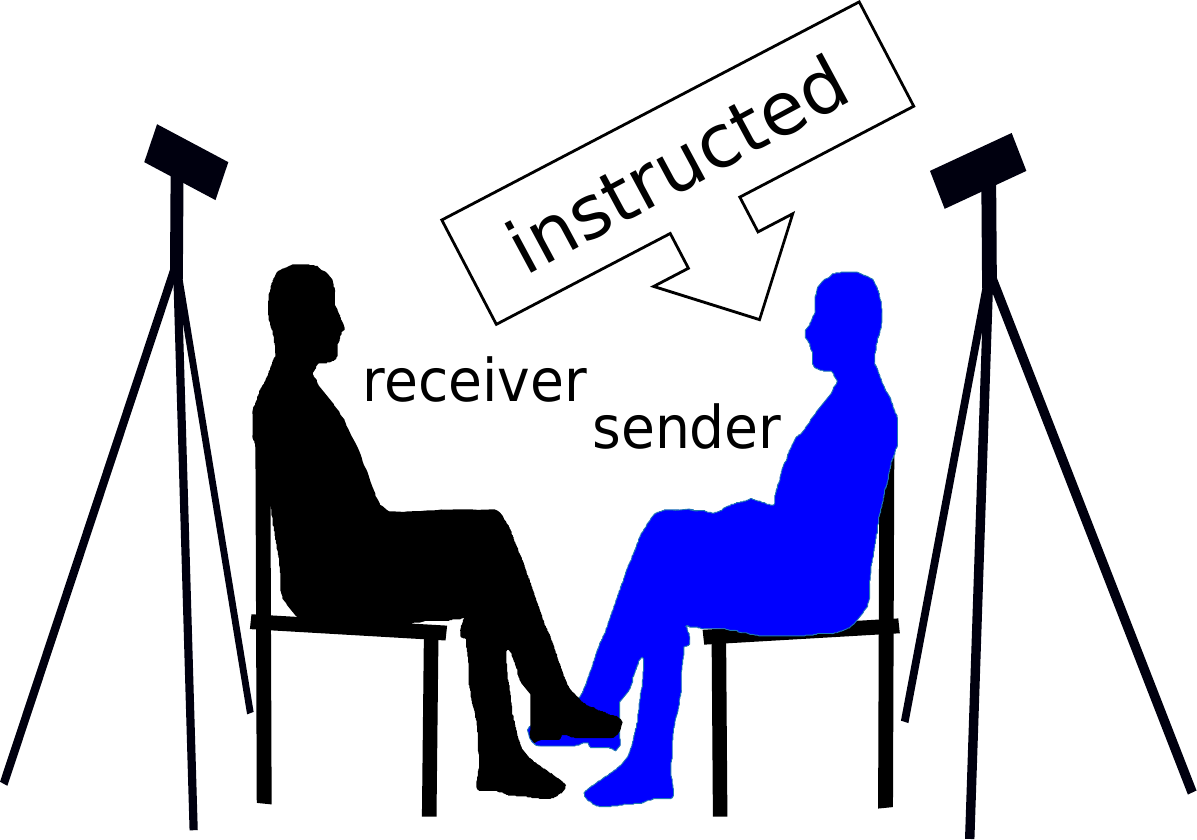}
 \caption{Experimental setup showing an example interaction pair (i.e, the sender and receiver) sitting face to face with camera positions towards each interacting partner's face. }
\label{fig:ExperimentalSetup}
\end{figure}

We created an experimental setup  (Figure \ref{fig:ExperimentalSetup})  in which two participants sat face to face while talking about their personal weaknesses. One participant was in the assigned role of the receiver (R), the other was in the assigned role of the sender (S).  S is instructed to take on a certain attitude (i.e., respectful, objective, contemptuous) whereas R is unaware of this and acts spontaneously.

In total, participants were asked to talk to each other about their personal weaknesses three times, either in circumstances of a respectful, contemptuous, or objective/neutral situation. Both partners were given specific times when to speak and when to listen. In all three experimental conditions, each participant kept their initially assigned role of acting as a S or R. The experimental conditions (i.e., respectful, contemptuous, objective) were conducted in various orders for all pairs to avoid the possible confounding effect of the order of these conditions. The objective/neutral condition consisted of the instruction that S and R should behave as neutral as possible (i.e., trying not to react readily towards the interacting partner). Instead, they should take their time to consider the world through the eyes of the other in order to understand their perspective in an uninvolved manner. As only S had the active experimental interaction attitude task (i.e., to behave either respectfully, contemptuously, or objectively), we
expected S to influence R in relevant facial expressions. In order to avoid flirtatious situations, that may overwrite the instructed condition, interacting partners were always from the same gender. See materials for the experimental instructions in the supplementary materials, part A.

Consistency of emotional influence was secured by the fact that after each conversation both interacting partners filled in a questionnaire in which they indicated in a self-report manner their positive affect during the interaction (i.e., on 4 items using a 7-point Likert scale). This measure revealed a clear linear function: In the respectful condition both interacting  partner indicated the highest positive affect (mean value: M= 5.61  and standard deviation: SD= 0.91), followed by the objective condition (M = 5.40, SD = .98), followed by the contemptuous condition (M= 4.15, SD= 1.36).

To capture nonverbal facial behavior, we positioned two frontal perspective cameras (Figure \ref{fig:ExperimentalSetup}), recording at 25 frames per second. Camera positions  and lighting conditions were optimized during a test session before the study started. This ensured high video quality in terms of a plain frontal view of the faces and  two-sided illumination. Motion blur rarely occurred, but could not be prevented entirely, especially in cases of faster movements like head turns. 
Except for the label of the experimental condition no other information (e.g., expression annotation per frame) were available for image analysis. The entire data set consisted of 34 pairs (mean age = 20.72, 24 female pairs, German-speaking participants), three conditions per pair and about four minutes of video per condition for each of the interacting partners, thus about 13 hours of video material. All participants gave written informed consent. The study was conducted in accordance with the Declaration of Helsinki and approved by the Ethics Committee of the Friedrich Schiller University of Jena. The ethical clearance number is FSV 18/04. Further, the study was pre-registered with aspredicted.org under the title ”Social Communication” $\#$13078 (See pre-registration in the supplementary materials, part B). 

The psychological research question was, whether and how  S  and  R  influence each other given different attitude situations. First, we expected more harmonic expressions (i.e., happiness) when both interacting partners were confronted with medium to high levels of respect (i.e., respectful and objective/neutral vs. contemptuous). Further, we expected the strongest activation of negative expressions (e.g., sadness) in the disrespectful condition (i.e., contemptuous vs. respectful and objective/neutral). Last, we expected, for all emotional facial actions,   S   to cause the effects and influence  R . In terms of the different facial expressions, we expected the strongest GC causality from  S  to  R   for positive expressions (i.e., happiness), followed by negative expressions (e.g., sadness).

\section{Methodology}
\label{sec:methodology}
\begin{figure}[!th]
\centering
\includegraphics[width=0.5\textwidth]{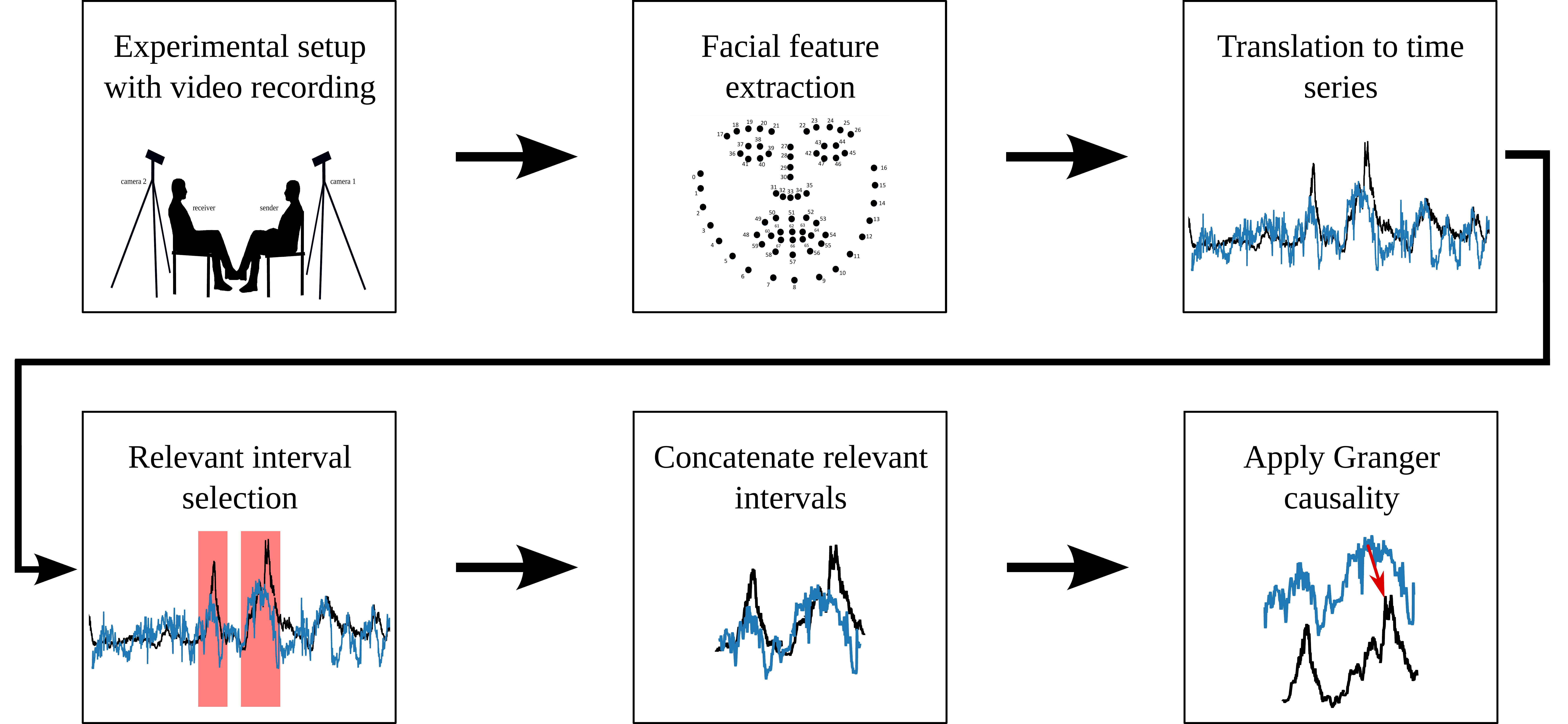}\caption{
The workflow of the  proposed concept for analyzing the direction of emotional influence  in dyadic dialogues.}
\label{fig:workflow} 
\end{figure}

  Figure \ref{fig:workflow} shows the workflow of the proposed method. In the following subsections, we introduce the feature extraction procedure,  the relevant interval selection approach, and the causal inference method.  Then, we combine all these steps to elucidate our entire approach.

\subsection{Facial Feature Extraction}
\label{sec:FeatureExtraction}
\noindent According to Ekman and Rosenberg \cite{ekman1997face}, facial expressions are the most important nonverbal signals when it comes to human interaction. The Facial Action Coding System (FACS) was developed by Ekman and Friesen \cite{ekman1978facial,ekman2002facial}. It  specifies facial  AUs, based on facial muscle activation. Examples of  AUs are the \textit{inner brow raiser}, the \textit{nose wrinkler}, or the \textit{lip corner puller}. Any facial expression is a combination of facial muscles being  activated, and thus, can be described by a combination of  AUs. Hence, the six  basic emotions (\textit{anger}, \textit{fear}, \textit{sadness}, \textit{disgust}, \textit{surprise}, and \textit{happiness}) can also be  represented via  AUs.  Thus, when for example  AU 6 (\textit{cheek raiser}), 12 (\textit{lip corner puller}), and 25 
(\textit{lips part}) are activated the facial expression \textit{happiness} is visible \cite{langner2010presentation}.

In general, all facial dynamic expressions are visual nonverbal communication cues transferable to time-series. Regarding our real experimental data, this approach is reasonable for positive emotions like \textit{happiness}, which is frequently visible throughout the dyadic interactions. Yet, it is not applicable for negatively associated emotions such as \textit{anger, disgust, fear,} or \textit{sadness}, as these emotions tend to be rare in dyadic interactions and thus are often only visible in subtle manners.  This is emphasized in Table \ref{tab:emotionCounts}, where we compute the visibility of emotions as defined in \cite{langner2010presentation} in all  recordings.

\begin{table}[!h]
\centering
\caption{Percentage of frames where emotions were visible  across all experimental conditions.}

\begin{tabular}{|c|c|}
  \hline
  Emotion & Occurrence (in \%) \\
  \hline
  Happiness & 12.9 \\
  Surprise & 1.00 \\
  Anger & .15 \\
  Disgust & 3.07\\
  Fear & .06 \\
  Sadness & 1.55 \\
  \hline
\end{tabular}
\label{tab:emotionCounts}
\end{table}
Wegrzyn et al. \cite{wegrzyn2017mapping} studied the relevance of facial areas for emotion classification and found differences in the importance of the eye and mouth regions.  AUs can be divided into upper and lower  AUs \cite{cohn2007observer}. Upper  AUs belong to the upper half of the face and cover the eye region, whereas  AUs in the lower face half cover the mouth region. We decided to split emotions into upper and lower emotions, 
according to the affiliation of  AUs to upper and lower face regions. For example, instead of using \textit{sadness} as a combination of  AU1,  AU4,  AU15 and  AU17 we used \textit{sadness upper} (AU1 and  AU4) and \textit{sadness lower} (AU15 and  AU17). 
All other emotions were split according to their  AUs belonging to the upper or lower facial half (Table \ref{tab:expressionAndAUs}). This procedure ensured that subtle facial expressions were also detectable.

\begin{table}[!h]
\centering
\caption{Expressions and the corresponding  AUs.}

\label{tab:expressionAndAUs} 
\begin{tabular}{|c|c|}
\hline
Expression            & Active AUs   \\
\hline
Happiness upper & 6        \\
Happiness lower & 12, 25        \\
Surprise upper & 1, 2, 5     \\
Surprise lower & 26           \\
Disgust lower  & 9, 10, 25    \\
Fear upper     & 1, 2, 4, 5   \\
Fear lower     & 20, 25       \\
Sadness upper  & 1, 4         \\
Sadness lower  & 15, 17       \\
Anger upper    & 4, 5, 7      \\
Anger lower   & 17, 23, 24  \\
\hline
\end{tabular}
\end{table}

\begin{figure}[!th]
\centering
\includegraphics[width=0.4\textwidth]{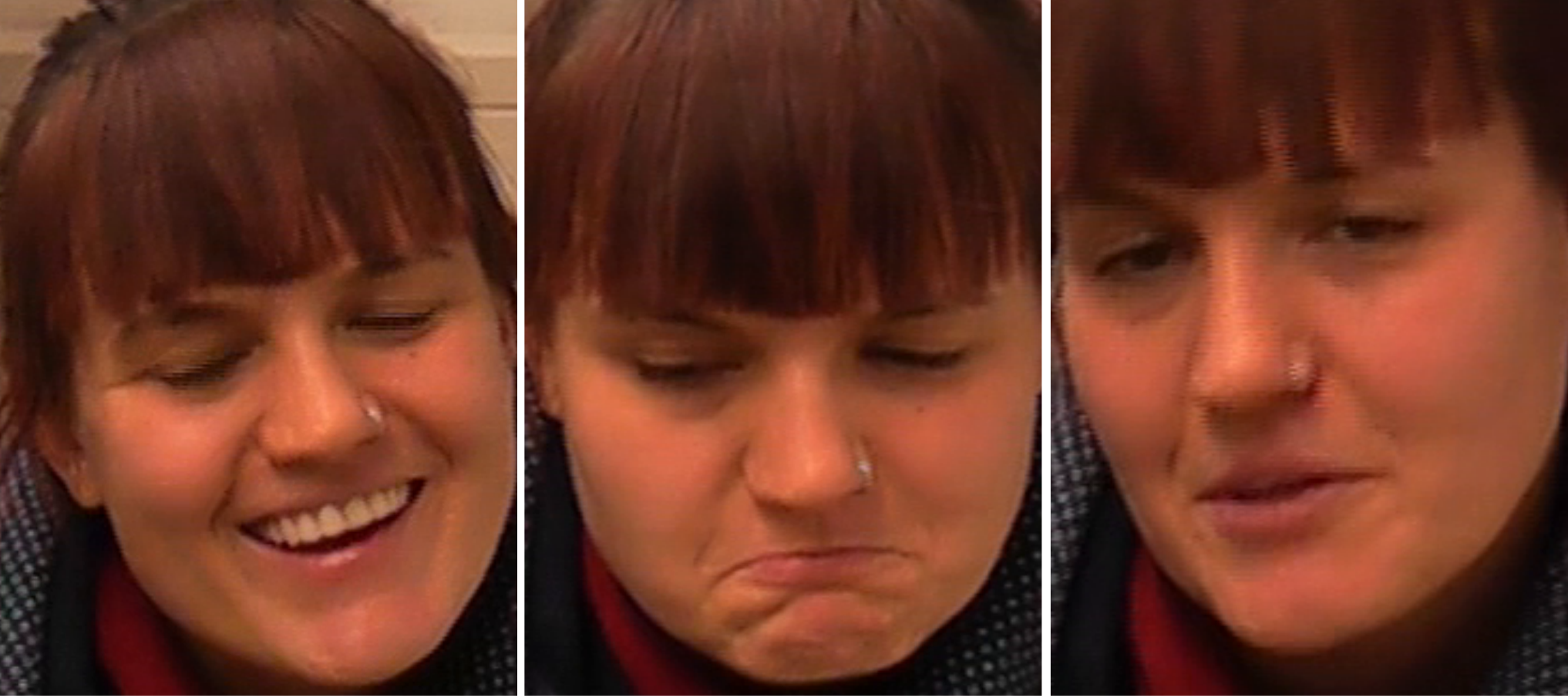}
\caption{ Participant with different facial expressions. From left to right: \textit{happiness lower}, \textit{sadness lower}, and \textit{sadness upper}}.  
\label{fig:AUSActivation}
\end{figure}

In Table \ref{tab:emotionCountsUpperLower} the detection percentage of upper and lower 
expressions is illustrated. After splitting, the emotional expressions \textit{anger lower}, \textit{sadness lower}, 
\textit{sadness upper}, and \textit{surprise lower}   could be detected in over 7\% of  the video material on average. 
Figure \ref{fig:AUSActivation} illustrates  participant with different facial expressions.

\begin{table}[!ht]
\centering
\caption{Percentage of  emotional expressions in the upper and lower face parts visible throughout 
experiment.}
\label{tab:emotionCountsUpperLower}
\begin{tabular}{|c|c|}
  \hline
  Emotion & Detection (in \%) \\
  \hline
  \textit{ Anger lower } & 8.20 \\
  \textit{ Anger upper } & .67 \\
  \hline
  \textit{ Disgust lower } & 3.07 \\
  \hline
  \textit{ Fear lower } & 5.94 \\
  \textit{ Fear upper } & .97 \\
  \hline
  \textit{ Happiness lower } & 16.24 \\
  \textit{ Happiness upper } & 26.32 \\
  \hline
  \textit{ Sadness lower } & 9.28 \\
  \textit{ Sadness upper } & 7.05 \\
  \hline
  \textit{ Surprise lower } & 26.91 \\
  \textit{ Surprise upper } & 2.48 \\
  \hline
\end{tabular}
\end{table}

\begin{figure}[!th]
\centering
\includegraphics[width=0.4\textwidth,height=3cm,keepaspectratio]{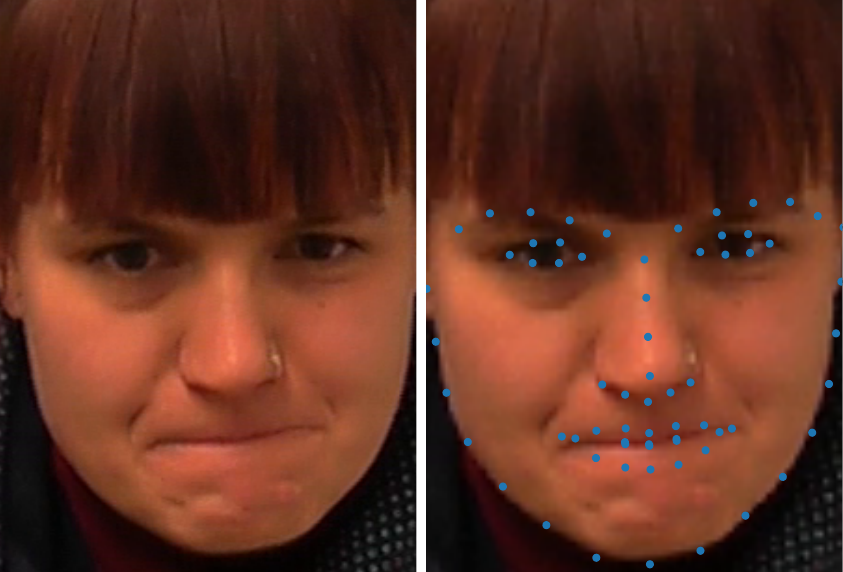}\\
(a)\\
\includegraphics[width=0.38\textwidth]{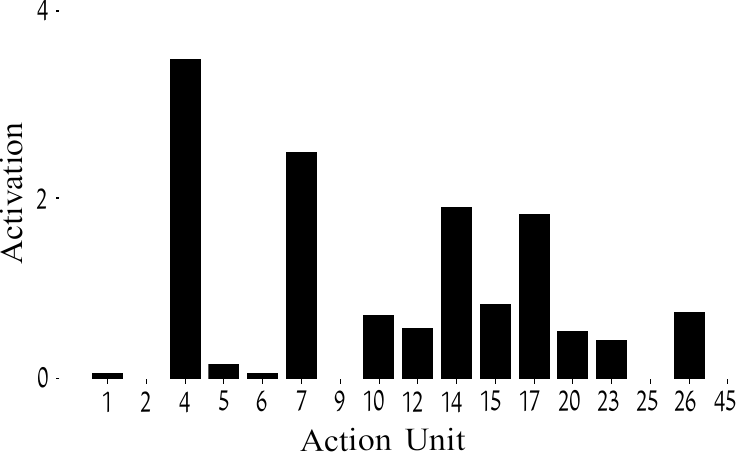}\\
\label{fig:AUDetectionOnly}
(b)\\
\caption{ Detection of facial expression \textit{angry}. (a) Landmarks of the  facial expression;  (b) Detected    AUs  by 
OpenFace: Strong activation of  AU4 (\textit{brow lowerer}), 7 (\textit{lip  
tightener}), 14 (\textit{dimpler}), and 17 (\textit{chin 
raiser})}
\label{fig:AUDetection}
\end{figure}

We evaluated the detection accuracy by mapping  AUs to emotions based on the AffectNet dataset \cite{mollahosseini2017affectnet}. We reached an accuracy of $25.8 \%$ for the six basic emotions plus contempt plus neutral. Especially happiness was detected very well, with $77\%$ accuracy, whereas fear and surprise classification seem to be more complex. For feature extraction, we used OpenFace 2.0 \cite{baltrusaitis2018openface,baltruvsaitis2015cross} which is a state-of-the-art,  open-source tool for  landmark detection; it estimates  AUs based on landmark positions. OpenFace preserves much of the information by regressing AUs instead of only classifying them and  is  capable of extracting 17 different  AUs (1, 2, 4, 5, 6, 7, 9, 10, 12, 14, 15, 17, 20, 23, 25, 26, 45) with an intensity scaled from 0 to 5. Figure  \ref{fig:AUDetection} illustrates the detection of landmarks and  AUs for an example image.

Let AUI  denotes the action unit  I, e.g., AU15 is action unit 15. From the OpenFace detections,  we compute each participant's  average face across the three conditions. That is, for or each participant we compute the mean  $\text{AUI}_{mean}$ and standard deviation  $\text{AUI}_{std}$ of the activation of $\text{AUI}$ over all three experimental conditions. We then count $\text{AUI}$ as activated in time frame k, if $\text{AUI}_{k} \ge \text{AUI}_{mean} + 0.5 * \text{AUI}_{std}$.  Then an  expression is counted as activated when all its corresponding AUs are activated. For example, \textit{sadness lower} is considered to be visible in frame $k$ if $\text{AU15}_{k} \geq \text{AU15}_{mean} + .5*\text{AU15}_{std}$ and $\text{AU17}_{k} \geq \text{AU17}_{mean} + .5*\text{AU17}_{std}$ hold. Following this step, the number of activations per person and per expression was counted for each  experimental condition, and normalized by the video length and maximum count of the expression.

\subsection{Causal Inference with Granger Causality}
\label{sec:GrangerCausality}

Let $X$ and  $Y$   be two stationary time series with zero mean and length $L$.  It is said that a time series   $X$  Granger causes a time series  $Y$ if the inclusion of past observations of  $X$ beside  $Y$ improves the prediction of   $Y$  significantly when compared to the prediction using only past values of $Y$. These two time series can be represented by the following two vector autoregressive  models.
\begin{equation} \label{eq:YgcX}
 X_t = \sum_{j=1}^M a_{j} X_{t-j} + \sum_{j=1}^M 
b_{j} Y_{t-j} + \varepsilon_t
\end{equation}

\begin{equation} \label{eq:XgcY}
Y_t = \sum_{j=1}^M c_{j} X_{t-j} + \sum_{j=1}^M 
d_{j} Y_{t-j} + \vartheta_t
\end{equation}
\noindent  where   $\quad t = 1 \dots L$  denotes the time index,  $\varepsilon_t$ and $\vartheta_t$ being two independent noise 
processes. The model order $M$  defines the maximum lag used to estimate  causal interactions. It can be  estimated using either Akaike \cite{Akaike1974} or  Bayesian Criterion \cite{Schwarz1978}.  The  model parameters  $a_j,b_j,c_j,d_j,   j=1, \ldots, M $ can then be   estimated using, for example,  the method of Least Squares (LS) \cite{Haykin1996}.

To test whether $Y$ Granger causes  $X$, two vector autoregressive  models are compared. The first  model, in which $Y$ is included for predicting $X$, as in  (\ref{eq:YgcX}). The second model is   
\begin{equation} \label{eq:XsimpleModel}
 X_{t} = \sum_{j=1}^M a'_{j} X_{t-j} + \varepsilon'_t
\end{equation}
where $Y$ is not included. Those models are then compared against each other via a statistical 
significance test, where the null hypothesis $H_0$ is tested against the 
alternative  $H_1$, with 
\begin{eqnarray}
H_0: b_1 = \dots = b_M = 0, \\
H_1: \exists b_k \neq 0  \quad k \in \{1 \dots M\}.
\end{eqnarray}
\noindent These hypotheses are equal to the variable selection problem in linear 
regression. Hence,  an  F-Test is applicable with 
\begin{equation}
 F = \frac{(|\Sigma_{\varepsilon'}| - |\Sigma_\varepsilon|)(T-2M-1)}{|\Sigma_{\varepsilon'}|~M} 
\end{equation}
where  $\Sigma_{\varepsilon'}$  is  the covariance matrix of the residual $\varepsilon_t'$ of the simple  model in (\ref{eq:XsimpleModel}),  and $\Sigma_\varepsilon$  is  the covariance matrix of the residual $\varepsilon_t$ of the  enriched  model in (\ref{eq:YgcX}) . $F$ follows an F-distribution, with $(M, T-2M-1)$ degrees of freedom. The  null hypothesis ($Y$ does not Granger cause  $X$) can be rejected at a level of significance $\alpha$, if $ F > F_{1-\alpha}(x;M,T-2M-1)$ where  $F_{1-\alpha}(x; M,T-2M- 1)$   denotes the value $x$, where the $F(x; M,T- 2M-1) = 1-\alpha$.

\noindent To test whether  $X$ Granger causes  $Y$, the above steps can be applied, but with the first model as in (\ref{eq:XgcY}), the second model being $Y_t =  \sum_{j=1}^M 
d'_{j} Y_{t-j} + \vartheta'_t$.  When testing for  GC , three different cases regarding the direction of  influence can occur \cite{schulze2004granger}:
\begin{enumerate}
 \item[1.] If $c_k = 0$ for $k = 1 \dots M$ and $\exists b_k \neq 0$ for $1 \leq k 
\leq M$ then  $Y$ Granger causes  $X$.
\item[2.] If $b_k = 0$ for $k = 1 \dots M$ and $\exists c_k \neq 0$ for $1 \leq k 
\leq M$ then  $X$ Granger causes  $Y$.
\item[3.] If for both $\exists b_k \neq 0$ for $1 \leq k \leq M$ and $\exists c_k \neq 
0$ for $1 \leq k \leq M$ then a bidirectional (feedback) relation exists.
\end{enumerate}

\noindent If none of the above cases holds,  $X$ and  $Y$ are not Granger causing each other.

\subsection{Relevant Interval Selection }
\label{sec:InervalSelection}
\noindent Considering the experimental setup, we had to expect multiple temporal scenes, further referred to as subintervals, in which the participants influenced each other. The time spans where causality is visible might range from half a second to half a minute. This may occur several times and can be interrupted by irrelevant scenes that differ in the length of time. As outlined above, the direction of influence in a subinterval can either be bidirectional or unidirectional driven by either  S  or  R . This implies that three unwanted effects can occur if the full-time span is analyzed.  First, temporal relations are not found at all; second, bidirectional relations mask temporal 
unidirectional relations and; third, unidirectional relations from  $X$ to 
 $Y$ mask temporal bidirectional influences or unidirectional influences from 
 $Y$ to  $X$ and vice versa.

\begin{figure}[!h]
\centering
\includegraphics[width=0.45 \textwidth]{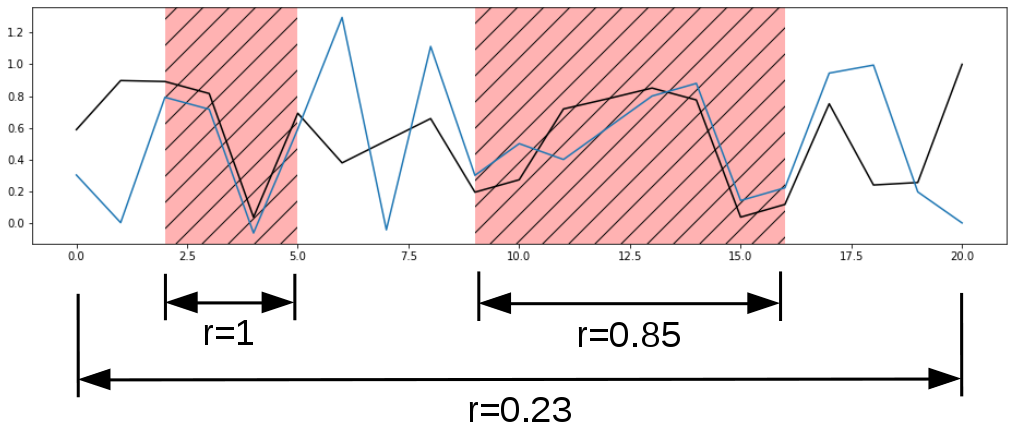}
\includegraphics[width=0.45\textwidth]
{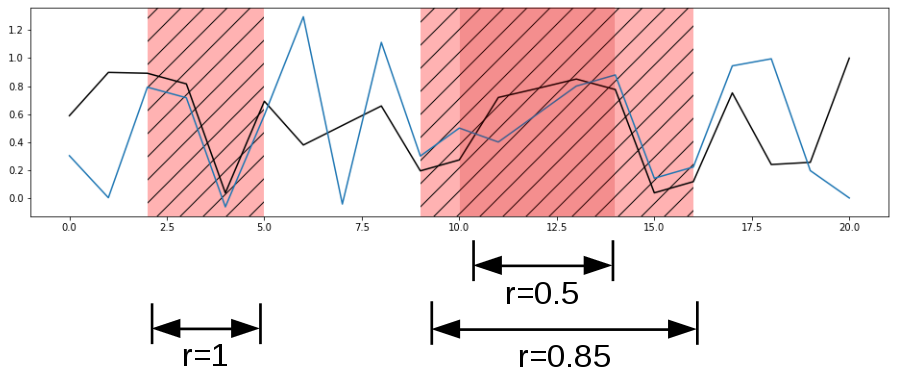}
\caption{Synthetic example of two time series showing wrong reasoning of correlation due to surrounding intervals. Upper graph shows an example of highly correlated subintervals of these time series (Pearson correlation parameter,  r=0.85 and r=1) within low correlated interval (r=0.23). While the lower graph shows an example of weakly  correlated subinterval (r=0.5) within highly correlated interval (r= 0.85).}
\label{fig:corsubints}
\end{figure}

The estimation accuracy of the  causal effect intensity  parameters $a_j, b_j, c_j, d_j, j=1, \dots,  M$  is mainly influenced by the accuracy of  estimating the correlation  of the two time series  $X$ and   $Y$  at different time shifts. When the time series contain  several intervals of irrelevant information, the transient similarity in  $X$ and   $Y$   may not be captured. Figure \ref{fig:corsubints}  illustrates that two-time series can have  highly correlated  subintervals within low correlated full time span interval   or   low correlated subintervals embedded in high correlated full time span interval. That is to say, that similarity measures that are applied to the entire time range would fail to capture transient similarities.  

Our central idea is to apply  GC  only to time series obtained by concatenating highly coherent (e.g., in terms of Pearson correlation) subintervals of raw time series. Instead of using a brute force algorithm, we suggest using a bottom-up approach for finding the longest set of maximal, non-overlapping, correlated intervals in time-series as proposed by Atluri et al.  \cite{atluri2014discovering1}. The authors applied their approach to fMRI data where they achieved good results for clustering coherent working brain regions.

Let  $X$ and  $Y$ be two time series of length $L$. An interval is called \textit{correlated interval} for a threshold $\beta$ when all its subintervals up to a lower interval length $L_{min}$ are correlated as well. An 
interval $I_{(a,b)}$ from $a$ to $b$ is called maximal, when $I_{(a,b)}$ is a \textit{correlated interval}, but $I_{(a-1, b)}$ and $I_{(a,b+1)}$ are not. And two intervals $I_{(a,b)}$ and $I_{(c,d)}$ are called non-overlapping, when $I_{(a,b)} \cap I_{(c,d)} = \emptyset$. From all intervals fulfilling these conditions the longest set (total length of intervals) is computed.

In the multivariate case, e.g.,   when multiple  AUs define an expression, we propose to compute the set of relevant intervals for each AU. Using the detection
of each  AU pair, the intersection over all AUs can be used as the set of selected relevant intervals. In the following, the set of selected relevant intervals between two time-series  $X$ and  $Y$ is called $AW_{XY}$. 

  \begin{figure}[!h]
   
 \includegraphics[width=0.5\textwidth]{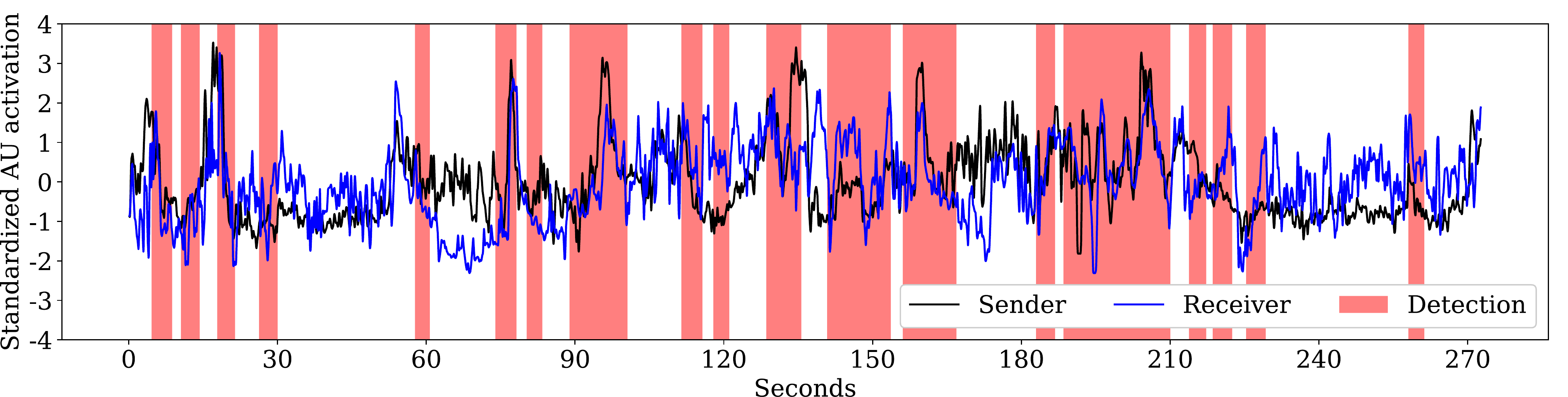}\\
 {\small (a) Relevant interval selection for AU6.}\\
    \par \par   
    \includegraphics[width=0.5\textwidth]{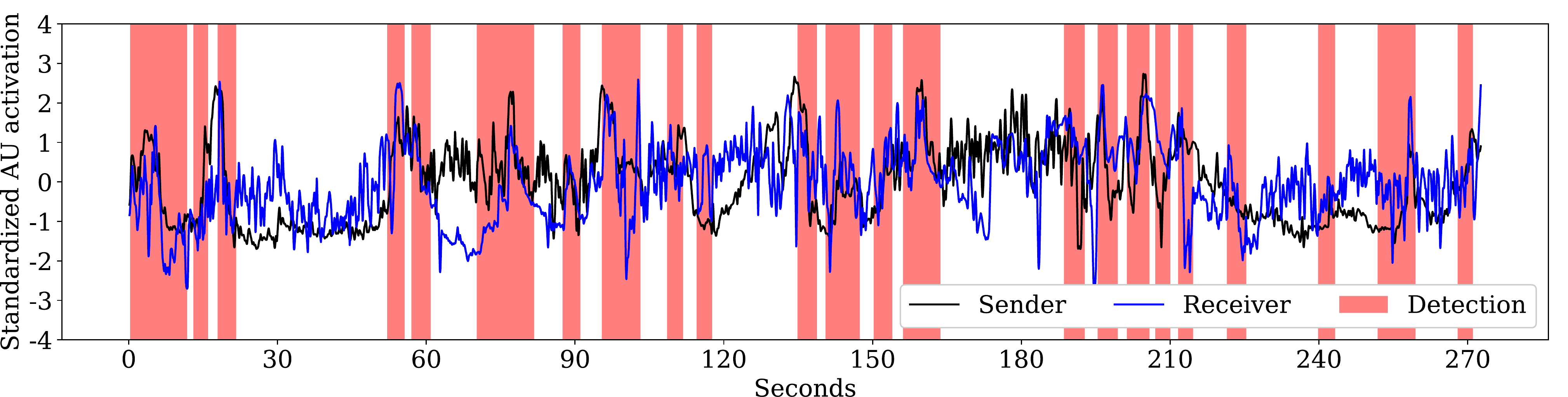}\\
       {\small (b) Relevant interval selection  for  AU12.}\\
      \par \par
 \includegraphics[width=0.5\textwidth]{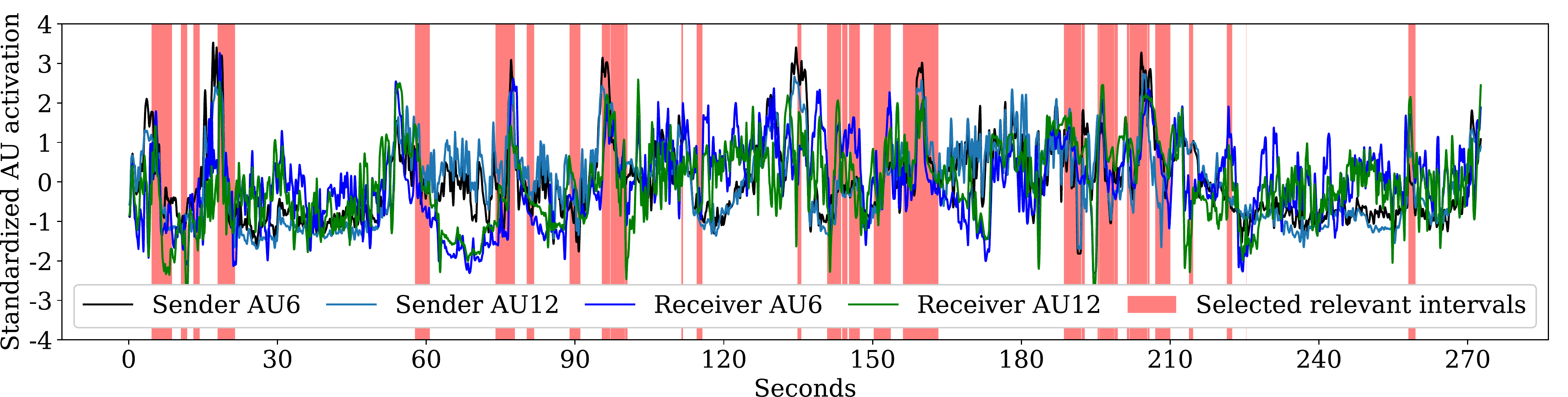}\\
  {\small (c) Selected  intervals for a facial expression with  AU6 and  AU12 being activated ((a) $\cap$ (b)). }\\
\par\par
 \includegraphics[width=0.5\textwidth]{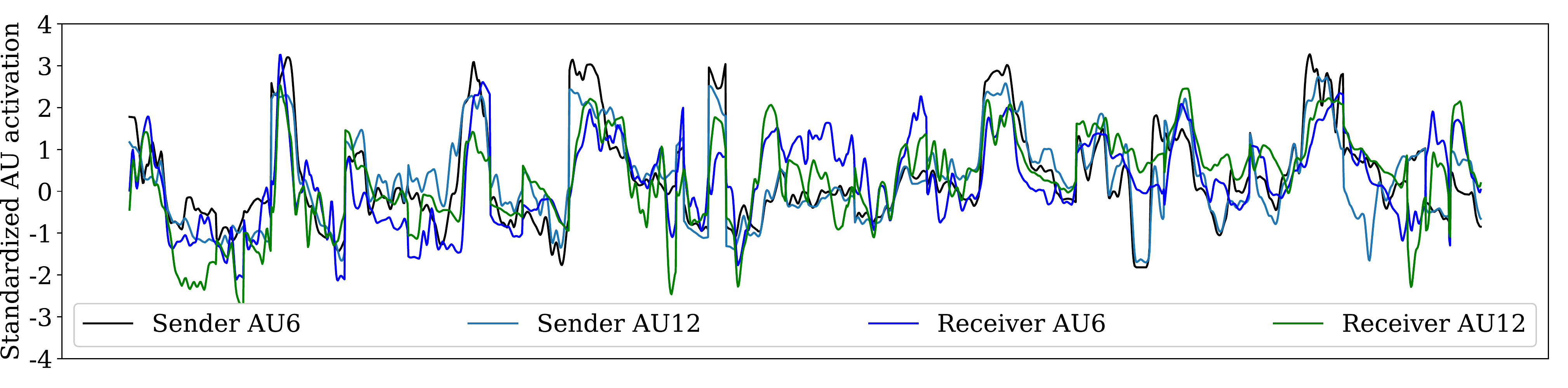}
 {\small (d)  Concatenation of the intervals  highlighted  in (c).}\\
\caption{Process of relevant interval selection in multivariate case, exemplified by  AU6 and  AU12.} 
\label{fig:RIS}
  \end{figure}


Figure \ref{fig:RIS} illustrates the selection process, exemplified by  AU6 and  AU12. First, the relevant intervals are computed for time 
shifts $s \in \{0,4,8,12\}$ and minimum interval length $L_{min}=75$, between 
 S  and  R, for  AU6 and  AU12 separately. After that step, 
additional processing steps, such as median filtering can be applied. 
Figure \ref{fig:RIS} (a) shows the relevant intervals of  AU6 and 
Figure \ref{fig:RIS} (b) shows the relevant intervals for  AU12, where both sets of relevant intervals were median filtered with kernel size 51. Thereafter, the selected relevant intervals are obtained by computing the intersection between the relevant intervals of  AU6 and  AU12, as visualized in Figure 
\ref{fig:RIS} (c). Finally, the selected relevant intervals can be concatenated for further processing, as illustrated in Figure 
\ref{fig:RIS} (d).

\subsection{Causal Inference with  Relevant Interval Selection and Granger Causality}
\label{ref:ModelCauseEffectRelations}
\noindent 
There are two major challenges in the analysis of the emotional causal effect relation in dyadic dialogues. First, due to the constructed situations, strong distinct emotions, computed by using traditional  AU combinations, were barely visible. Second, the time-variant and situation-dependent communication, resulted in high variety and volatility of time spans in which causal effect behavior between interacting partners is visible. To tackle these difficulties, we use a combination of the time series of the  facial features described  in  (\ref{sec:FeatureExtraction}) and the proposed relevant interval selection approach (\ref{sec:InervalSelection}) as detailed in the following steps. 

\begin{itemize}
\item[1.]  We applied the relevant interval selection approach pairwise to the time series of the identified  AUs of all of  the relevant facial expression as illustrated in Figure \ref{fig:RIS}, with a minimum interval length of 75 frames and a threshold of 0.8 for Pearson correlation. Based on known average human reaction times (ca. 200 ms or 6 frames \cite{jain2015comparative}), we shifted one time series by 0, 4, 8, and 12  frames both, back and forth in time, and computed relevant intervals. The grid selected for shifting does cover quicker and slower reactions of participants.
    Afterwards, we computed the longest set of the list of relevant intervals obtained from the different shifts. Before computing  GC, we median filtered the selected intervals with a filter length of 51 (2 seconds) and extended the intervals by 12 frames on each side. We removed frames for both, S and R, when for either S or R the OpenFace confidence value was below .89.  The confidence score gives a rough orientation how reliable the AU score computed by OpenFace is. A low confidence score might occur when, for example,  the face is occluded or the person moved quickly, which can lead to blurred images. 	
\item[2.]  We calculated the average  GC  on  the set of selected intervals of the  standardized time-series. The results were counted according to the possible outcomes of the  GC  test as described in Section \ref{sec:GrangerCausality}, as either unidirectional caused by  S, unidirectional caused by  R, bidirectional, or no causality. 
\end{itemize}

\section{Experimental Results and Discussion}
\label{sec:results}

\subsection{Appearance  of facial expressions in the different experimental  conditions}  
In order to compare the experimental conditions (i.e., respectfully, contemptuously, and objectively) with regards to total counts of facial expressions for all participants (i.e., S and R),  we applied a Wilcoxon signed-rank test. This allowed us to compare how often a specific expression was visible over the full-time span between the different experimental conditions. For testing, we used a Benjamini-Hochberg p-value correction \cite{benjamini1995controlling} with a false discovery rate of $Q = .05$ and individual p-values of 0.05.  Table \ref{tab:signed-rank} summarizes the results of the  Wilcoxon signed-rank test. As expected,  participants showed significantly more happiness upper and lower in the respectful condition than in the contempt and objective condition. Furthermore, as expected, we found more sadness lower and upper expressions in the contempt compared to the objective condition. This is fully in line with our psychological hypotheses and demonstrates that our instructed attitude manipulations for the sender had an effective influence on both interaction partners (i.e., sender and receiver). Note, found expressions for fear lower and disgust lower occurred against expectations (i.e., those were found significantly more in the respectful compared to the objective condition). This is highly valuable information as it suggests that facial happiness and sadness expressions are the most indicative when comparing negative and positive interaction attitudes. Fear and disgust expressions seem to rather occur for attitudes between the extremes (i.e., respectful and contemptuous).  

\begin{table}[!h]
\caption{Results of the Wilcoxon signed-rank (W) test. The table shows only all expressions that were significantly (p-value $<0.05$) different in their occurrence counts between different conditions (Insignificant results are not shown). We permuted all experimental conditions and tested all expressions.}
\label{tab:signed-rank} 
\begin{tabular}{|l|c|c|c|c|}
\hline
Expression      & First   & Second  & p-value &  W test \\ 
								&   condition           &   condition         &          &   statistics           \\
\hline
Happiness upper & respectful      & contempt         & .0035   & 695             \\ 
                & respectful      & objective        & .0001 & 511             \\ \hline
Happiness lower & respectful      & contempt         & .0003 & 580             \\ 
                & respectful      & objective        & .0017  & 660             \\ \hline
Sadness lower   & contempt        & objective        & .0088   & 744             \\ 
                & respectful      & objective        & .0005  & 606             \\ \hline
Sadness upper   & contempt        & objective        & .001   & 635             \\ \hline
Fear lower      & respectful      & objective        & .007   & 730             \\ \hline
Disgust lower   & respectful      & objective        & .018   & 783             \\ \hline
\end{tabular}
\end{table}

\begin{table*}
\centering
\caption{Number  of pairs for which the Granger causality (GC) test, with p-value =0.05, showed a specific direction of influence in the respectful  condition. Average count is the average count of pairs  across all expressions. Dominant causal direction (the higher value between Sender GC Receiver (S GC R)  and  Reciever GC Sender (R GC S))  is shown in bold font.}
\label{tab1} 
\begin{tabular}{|c|c|c|c|c|c|c|c|c|}
\hline \hline
 {\multirow{2}{*}{Expression}} & \multicolumn{4}{c|} {Full Time Span }  & \multicolumn{4}{c|} {Relevant Interval Selection } \\  \cline{2-9}
   & S GC R &  R GC S & Bidirectional& No causality & S GC R & R GC S &  Bidirectional& No causality\\ \hline
Happiness lower & ${\bf 4}$ & 1 & 22 & 7  & ${\bf 8}$ & 4 & 14 & 8 \\
Happiness upper & 1 & ${\bf 3}$ & 24 & 6  & ${\bf 6}$  & 4 & 18 & 6 \\
Sadness lower   & 6 & 6 & 4  & 18 & ${\bf 9}$  & 3 & 8  & 14 \\
Sadness upper   & ${\bf 6}$ & 0 & 7  & 19 & $ 7$  & {\bf 8} & 6  & 11 \\
\hline 
Average count & ${\bf  4.25}$    & 2.5    & 14.25         & 12.5    & ${\bf  7.5}$  & 4.75   & 11.5     & 9.75 \\
\hline \hline
\end{tabular}
\end{table*}

\begin{table*}
\centering
\caption{Number  of pairs for which the Granger causality (GC) test, with p-value =0.05, showed a specific direction of influence in the contempt  condition. Average count is the average count of pairs  across all expressions. Dominant causal direction (the higher value between Sender GC Receiver (S GC R)  and  Reciever GC Sender (R GC S))  is shown in bold font.}
\label{tab2} 
\begin{tabular}{|c|c|c|c|c|c|c|c|c|}
\hline \hline
 {\multirow{2}{*}{Expression}} & \multicolumn{4}{c|} {Full Time Span }  & \multicolumn{4}{c|} {Relevant Interval Selection } \\  \cline{2-9}
   & S GC R &  R GC S & Bidirectional& No causality & S GC R & R GC S &  Bidirectional& No causality\\ \hline
Happiness lower & ${\bf 3}$ & 2 & 17 & 12 & ${\bf 8}$ & 4 & 10 & 12  \\
Happiness upper & ${\bf 4}$ & 1 & 24 & 5  & 3 & ${\bf 7}$ & 17 & 7 \\
Sadness lower   & ${\bf 5}$ & 3 & 10 & 16 & 7 & ${\bf 8}$ & 1  & 18 \\
Sadness upper   & ${\bf 5}$ & 3 & 2  & 24 & 4 & ${\bf 5}$ & 6  & 19 \\
\hline \hline
Average count & ${\bf 4.25}$  & 2.25   & 13.25         & 14.25   & 5.5  & ${\bf 6}$      & 8.5      & 14   \\
\hline
\end{tabular}
\end{table*}

\begin{table*}
\centering
\caption{Number  of pairs for which the Granger causality (GC) test, with p-value =0.05, showed a specific direction of influence in the neutral/objective  condition. Average count is the average count of pairs  across all expressions. Dominant causal direction (the higher value between Sender GC Receiver (S GC R)  and  Reciever GC Sender (R GC S))  is shown in bold font.} 
\label{tab3} 
\begin{tabular}{|c|c|c|c|c|c|c|c|c|}
\hline \hline
 {\multirow{2}{*}{Expression}} & \multicolumn{4}{c|} {Full Time Span }  & \multicolumn{4}{c|} {Relevant Interval Selection } \\  \cline{2-9}
   & S GC R &  R GC S & Bidirectional& No causality & S GC R & R GC S &  Bidirectional& No causality\\ \hline
Happiness lower & 0 & ${\bf 5}$ & 23 & 4  & ${\bf 9}$  & 7  & 7 & 9 \\
Happiness upper & ${\bf 5}$ & 0 & 24 & 5  & ${\bf 9}$  & 4  & 21 & 0 \\
Sadness lower   & 2 & ${\bf 3}$ & 7  & 22 & ${\bf 6}$  & 7  & 9 & 12 \\
Sadness upper   & ${\bf 4}$ & 3 & 3  & 24 & ${\bf 10}$ & 5  & 2 & 17 \\
\hline \hline
Average count & 2.75   & 2.75   & 14.25         & 13.75   &${\bf 8.5}$   & 5.75    & 9.75     & 9.5   \\ \hline  
\hline
\end{tabular}
\end{table*}
\subsection{Direction of emotional influence }  
In order to study the direction of emotional influence, we compared results of  GC test on the relevant interval selection approach versus results of GC test on the full-time span approach. The comparison is represented by  the count of pairs for which the Granger causality (GC) test, with p-value =0.05, showed a specific direction of influence, under the three experimental conditions (Tables \ref{tab1} (i.e., respectful), \ref{tab2} (i.e., contemptuouse), and \ref{tab3} (i.e., objective)), for the expressions: happiness upper/lower and sadness upper/lower.   These results clearly indicate that the use of the relevant interval selection approach prior to causal inference resulted in considerably more pairs showing unidirectional causation as well as less bidirectional or no causation. 

Most interestingly, S influences R particularly in the respectful and objective/neutral compared to the contemptuous condition. In the contemptuous condition the pattern of dominant influence changes (i.e., on average R and  S influence each other similarly) when using relevant interval selection approach. Psychologically this indicates that the receiver in this condition is actively trying to repair the overall negative interaction quality, for example by inducing emphatic concern in the sender. These result indicate that using facial expressions only holds the potential to reveal covert attitudes and behaviors that would easily be missed and overlooked when working with verbal behavioral cues. That is most likely in situations like these (i.e., an interaction partner acting ignorant and dismissive),  a receiver of such contemptuous information would simply produce less speech content.  Finally we can notice that inline with prediction there is significantly higher bidirectional influence for positive emotions than in negative emotions in all experimental conditions. However, while we expected considerable reduction in emotional influence from S to R for negative expression when comparing to positive expression, our results indicate that there  is only a slight  reduction  in the influence from S to R  for negative expressions,  such as sadness lower and upper when using relevant interval selection.  For the full time span, the influence of S on R is even higher for negative emotions than positive ones which is against prediction. 
Overall these results indicate that the use of the relevant interval selection  approach significantly improved the detection of the emotional influence in all experimental  conditions.

\section{Conclusions}
\label{sec:conclusion}
In this paper, we have presented a complete concept for   identifying the direction of emotional  influence in nonverbal dyadic communication when starting with raw video materials using facial expressions only.   To this end, we presented an algorithm for the extraction of emotional facial features, capable of capturing emotional expressions even when strong distinct emotions are not visible.  To improve causal inference we proposed an intelligent interval selection approach for filtering relevant information in dyadic dialogues.  Subsequently, we were able to apply Granger causality  to the set of selected relevant intervals and compute the direction of influence. We applied our approach to real data obtained from a psychological experimental setup. The obtained results revealed that the use of the relevant interval selection  approach combined with the proposed facial features significantly improved the detection of the emotional influence for dyadic communication in various instructed interaction conditions. This work also allowed a major step forward in the 2nd person social sciences as we were able to study social emotions in a truly interactive manner. Further, the results of this study  indicate that using facial expressions only holds the potential to study implicit attitudes and behaviors in emotionally laden circumstances that would easily be missed when using speech-content information.  Overall, we identify our contribution as an important step towards an interdisciplinary research that combines computer vision potentials, psychological  observation, and theoretical knowledge of causality methods,  to gain novel insights into emotions in real-time social encounters.  For further research, we suggest using a  learning system capable of classifying upper and lower emotional expressions  based on all  AUs. Also it would be interesting to tear apart the contributions of various social information channels (i.e., speech, non-verbal speech, facial expressions) towards outcomes of the interaction experience (e.g., enjoyment, engagement, and liking of interacting partner).

\section*{ACKNOWLEDGMENTS}
The authors thank the Carl Zeiss Foundation for the financial support within
the scope of the program line "Breakthroughs: Exploring Intelligent Systems"
for "Digitization — explore the basics, use applications". D. Schneider was supported by the DFG Scientific Network "Understanding Others" - SCHN 1481/2-1. 

\bibliographystyle{ieeetr}

\bibliography{nonverbal}

\end{document}